\documentclass[11pt,a4paper]{llncs}
\usepackage{amsmath}
\usepackage{amssymb}
\usepackage{graphicx}
\usepackage{subcaption}
\usepackage{float}
\usepackage{marvosym}
\usepackage{url}
\usepackage{fancyhdr}
\usepackage{listings}
\usepackage{xcolor}

\usepackage{geometry}
\usepackage{hyperref}
\newcommand{\keywords}[1]{\par\addvspace\baselineskip
\noindent\keywordname\enspace\ignorespaces#1}

\fancypagestyle{firstpage}{\fancyhf{}
}

\begin{document}


\title{\LARGE{Analysing the Robustness of Vision-Language-Models to Common Corruptions}}


%
%
\author{\large{M. Usama*$^1$  \and S. A. Asim*$^1$ \and S. B. Ali*$^1$ \and S. T. Wasim$^2$ \and U. B. Mansoor$^1$}}
\institute{\large{$^1$Department of Electrical Engineering, DHA Suffa University \\ Karachi, Pakistan \\ $^2$Department of Computer Science, University of Bonn \\ Bonn, Germany}\\ \small{*Indicates equal contribution, authors in alphabetical order.}}

%


%
%


\maketitle

\thispagestyle{firstpage}

\begin{abstract}
Vision-language models (VLMs) have demonstrated impressive capabilities in understanding and reasoning about visual and textual content. However, their robustness to common image corruptions remains under-explored. In this work, we present the first comprehensive analysis of VLM robustness across 19 corruption types from the ImageNet-C benchmark, spanning four categories: noise, blur, weather, and digital distortions. We introduce two new benchmarks, TextVQA-C and GQA-C, to systematically evaluate how corruptions affect scene text understanding and object-based reasoning, respectively. Our analysis reveals that transformer-based VLMs exhibit distinct vulnerability patterns across tasks: text recognition deteriorates most severely under blur and snow corruptions, while object reasoning shows higher sensitivity to corruptions such as frost and impulse noise. We connect these observations to the frequency-domain characteristics of different corruptions, revealing how transformers' inherent bias toward low-frequency processing explains their differential robustness patterns. Our findings provide valuable insights for developing more corruption-robust vision-language models for real-world applications.
\end{abstract}
\section{Introduction}
\label{sec:intro}

Vision-Language Models (VLMs) have emerged as a powerful class of artificial intelligence systems capable of interpreting both visual and textual inputs. These models have demonstrated exceptional performance across various multimodal tasks, including image captioning, visual question answering, and image-text retrieval. However, real-world deployment of VLMs typically involves processing images with distortions, noise, and other corruptions introduced by environmental conditions, sensor limitations, or adversarial attacks. Despite their growing prominence, the robustness of VLMs under such challenging conditions remains an understudied research problem.

The human vision system exhibits robustness in ways that current computer vision systems do not, as demonstrated by the corruption benchmarks developed by Hendrycks and Dietterich~\cite{hendrycks2019benchmarking}. Unlike artificial vision systems, human vision is not easily misled by minor alterations in image presentation. The naked human eye remains reliable when confronted with various forms of corruption such as snow, blur, pixelation, or combinations of these effects.

Achieving comparable robustness represents a critical goal for computer vision and machine learning applications, particularly for deep learning systems intended for deployment in safety-critical contexts. While Hendrycks and Dietterich~\cite{hendrycks2019benchmarking} focused exclusively on vision systems, our study systematically analyzes the effects of common corruptions specifically on vision-language models. Although both vision systems and vision-language systems utilize similar vision encoders, we extend the analysis of Hendrycks and Dietterich~\cite{hendrycks2019benchmarking}, building upon their conclusion that years of architectural advancements yielded only minimal improvements in relative robustness. This underscores the importance of benchmarking and enhancing robustness in vision-language models.

Our study systematically examines the impact of common corruptions on VLM performance, particularly in visual question-answering (VQA) tasks. Using the TextVQA and GQA datasets as benchmarks, we apply controlled corruptions from the ImageNet-C framework across four categories: noise, blur, weather, and digital distortions (e.g., Gaussian noise, snow, pixelation, and motion blur), each at multiple severity levels. By evaluating the LLaVA 1.5 model's ability to generate accurate responses under varying degrees of image degradation, we identify key vulnerabilities in existing architectures. Our investigation addresses two primary research questions: (1) how different types and intensities of noise affect VLM accuracy, and (2) which architectural components contribute to robustness or susceptibility to corruption.

The findings from this research provide critical insights into the failure modes of modern VLMs and suggest potential architectural modifications to improve their resilience. By understanding how noise impacts feature extraction, attention mechanisms, and multimodal alignment, we can inform the development of more robust models for real-world applications. Ultimately, this work contributes to the broader goal of improving the reliability of AI systems in diverse and unpredictable environments.

Our main contributions in this paper include:
\begin{itemize}
   \item We provide the first systematic analysis of vision-language model robustness to common corruptions, contributing two comprehensive benchmarks: TextVQA-C and GQA-C, with a total of 190 evaluations (19 corruptions x 5 severity levels x 2 datasets).
   \item Our analysis of TextVQA-C reveals that scene text understanding is particularly vulnerable to blur corruptions and high-frequency noise patterns, with OCR capabilities degrading most severely under defocus blur and snow conditions.
   \item Through GQA-C, we demonstrate that object-based reasoning shows distinct degradation patterns, with spatial understanding being most robust to motion blur and brightness changes while deteriorating significantly under frost and impulse noise.
   \item We establish connections between corruption types and frequency domain characteristics, revealing how transformer-based VLMs exhibit a low-frequency processing bias that explains their differential robustness patterns across corruption categories.
\end{itemize}
\section{Related Work}
\label{sec:related}

This section reviews the foundational research areas relevant to our study on Vision-Language Model robustness. We begin by examining Contrastive Vision Language Encoders which provide the architectural backbone for multimodal understanding. We then explore Large Language Models (LLMs) and their evolution, followed by an overview of modern Vision-Language Models (VLMs) that build upon these advancements. Finally, we discuss research on model robustness, covering both naturally occurring corruptions and adversarial attacks that affect vision-language systems.

\subsection{Contrastive Vision Language Encoders}
Contrastive Vision Language Encoders are a powerful machine learning architectural element, providing the backbone of Vision and Language Models. CLIP (Contrastive Language–Image Pretraining)~\cite{radford2021learning} leverages Vision-language pre-training, aiming to learn multimodal foundation models with improved performance on various vision-and-language tasks. Depending on the downstream task, different model architectures have been proposed, including the dual-encoder architecture~\cite{radford2021learning}, the fusion-encoder architecture~\cite{tan2019lxmert}, the encoder-decoder architecture~\cite{zhu2022encoder}, and more recently, the unified transformer architecture~\cite{liu2022uninet}. Various pre-training objectives have also been proposed over the years, and have progressively converged to robust ones, such as image-text contrastive learning, as given in CLIP, image-text matching~\cite{yao2022image,radford2021learning}, and (masked) language modeling~\cite{zhou2021melm,yu2022point}. Most Vision-Language Pre-Training (VLP) methods perform end-to-end pre-training using large-scale image-text pair datasets. Some VLP methods freeze the image encoder, including the early work which adopts a frozen object detector to extract visual features~\cite{zheng2020ensemble}, and the recent LiT~\cite{zhai2022lit} which uses a frozen pre-trained image encoder for CLIP pre-training. Some methods freeze the language model to use the knowledge from LLMs for vision-to-language generation tasks~\cite{tsimpoukelli2021multimodal}. The introduction of BLIP-2~\cite{li2023blip} meant that models can now effectively and efficiently leverage both frozen image encoders and frozen LLMs for various vision-language tasks, achieving stronger performance at a lower computation cost. 

\subsection{Evolution of Large Language Models}
Large Language Models (LLMs) have experienced rising success in recent years due to the scaling up of training data and an increase in the number of parameters. Early models, such as BERT~\cite{devlin2019bert}, GPT-2~\cite{radford2019language}, and T5~\cite{raffel2020exploring}, provided the basis for this progress. BERT succeeded by enabling deep bidirectional understanding for NLP tasks, GPT-2 revolutionized text generation with large-scale auto-regressive modeling, and T5 unified all NLP tasks under a text-to-text framework for maximum flexibility. Subsequently, GPT-3~\cite{NEURIPS2020_1457c0d6}, with a massive scale of 175 billion parameters, was introduced, due to the success of its predecessor. This development inspired the creation of various other large language models, including MegatronTuring NLG~\cite{smith2022using}, Chinchilla~\cite{hoffmann2022training}, PaLM~\cite{chowdhery2023palm}, OPT~\cite{zhang2022opt}, and BLOOM~\cite{workshop2022bloom}. Additionally, Wei et al.~\cite{wei2022emergent} discovered several emergent abilities, which appear exclusively in large models under scaling. Moreover, by aligning the pre-trained large language model GPT-3 with human intent, instructions and human feedback, InstructGPT~\cite{ouyang2022training} and ChatGPT~\cite{openai2022chatgpt} enable conversational interactions with humans and can answer a wide range of diverse and complex questions. Advancing to LLM's such as Meta's LLaMA model~\cite{touvron2023llama}, more recently, several open-sourced models, such as Vicuna~\cite{vicuna2023}, have been developed based on LLaMA~\cite{touvron2023llama} and also exhibit similar performance.

\subsection{Modern Vision-Language Models}
Building upon the success of LLMs, Vision-Language Models (VLMs) such as LLaVA~\cite{liu2023llava} builds on the CLIP encoder and GPT-style language models to enable instruction-following in visual contexts. LLaVA is an end-to-end multimodal model designed to enhance vision-language instruction-following capabilities. It integrates a CLIP-based visual encoder with a transformer. Subsequently, MiniGPT-4~\cite{zhu2023minigpt} is a lightweight yet powerful vision-language model connected via a learned projection layer that aligns the visual and textual modalities. Pioneering studies like VisualGPT~\cite{chen2022visualgpt} and Frozen~\cite{tsimpoukelli2021multimodal}, in which the key challenge in using a frozen LLM is to align visual features to the text space. To achieve this, Frozen~\cite{tsimpoukelli2021multimodal} finetunes an image encoder whose outputs are directly used as soft prompts for the LLM. In contrast, Flamingo~\cite{alayrac2022flamingo} inserts new cross-attention layers into the LLM to inject visual features, and pre-trains the new layers on billions of image-text pairs. Both methods adopt the language modeling loss, where the language model generates texts conditioned on the image. have demonstrated the benefits of employing a pre-trained language model as a vision-language model decoder. Flamingo~\cite{alayrac2022flamingo} was then developed to align a pre-trained vision encoder and language model using gated cross-attention. Both methods, Flamingo, and Frozen adopt the language modeling loss, where the language model generates texts conditioned on the image. have demonstrated the benefits of employing a pre-trained language model as a vision-language model decoder. Most recently, PaLM-E~\cite{driess2023palm}, featuring 562 billion parameters, has been developed to integrate real-world continuous sensor modalities into an LLM, thereby establishing a connection between real-world perceptions and human languages. GPT-4~\cite{ai2023gpt} has also been recently released, showcasing more powerful visual under real-world scenarios and can accept image and text inputs and produce text outputs. Despite these innovations, VLMs are not inherently robust to perturbations. Real-world deployment requires assessing their vulnerability to visual corruptions and adversarial attacks.
\subsection{Robustness Challenges in Vision-Language Systems}
Vision-language systems are susceptible to naturally occurring corruptions such as weather, or corruptions brought about by the nature of the equipment, such as digital, or blur based transformations, as well as adversarial modifications such as noise. These can be divided into two categories, namely, naturally occuring corruptions, and adversarial attacks. ImageNet-C~\cite{hendrycks2019benchmarking} is a standard benchmark for evaluating image classifiers under 19 types of naturally occurring common corruptions at five severity levels, including noise, blur, weather, digital. While initially developed for image classification, ImageNet-C has been used in recent work to test vision system robustness under degraded visual input. Adversarial attacks, on the other hand, pose another challenge to robustness. These occur due to a bad actor, that induces an attack. These attacks can include: pixel-level attacks, patch-based backdoor attacks, sibling attacks. For pixel-level attacks, One-Pixel-Attack~\cite{su2019one}, and DeepFool~\cite{moosavi2016deepfool}, introduce imperceptible perturbations to fool the model. Similarly, to combat the vulnerabilities of neural networks towards adversarial attacks, Zhang et al. propose a two-stage ``separate-and-trace'' framework that embeds unique features into model copies using VAE-based training, enabling forensic tracing of adversarial examples back to the source model in buyer-seller settings~\cite{fang2023tracing}. Many defenses against adversarial attacks~\cite{frosio2023best} (eg robust classifiers, randomization, or image purification) use countermeasures put to work only after the attack has been crafted. Defending against patch-based backdoor attacks on self-supervised learning~\cite{tejankar2023defending}. Several machine learning models, particularly neural networks, are vulnerable to adversarial examples—small, worst-case perturbations that cause high-confidence misclassifications—due primarily to their linear nature, which explains their generalization across architectures and leads to a fast adversarial example generation method that improves robustness through adversarial training.~\cite{goodfellow2014explaining}. In contrast, Sibling-Attack is a novel black-box face recognition attack that enhances transferability by leveraging multi-task learning—specifically attribute recognition—to optimize cross-task gradients, significantly outperforming existing methods on both pre-trained and commercial FR systems~\cite{li2023sibling}. To further improve robustness, recent efforts involve adversarial training, contrastive learning under perturbations, and robust alignment techniques such as using masked image modeling or vision transformer enhancements to filter out noise-sensitive features. However, there is still a significant gap between clean and corrupted performance, motivating further investigation into the reliability of VLMs in real-world, noisy environments.
\section{Preliminaries: Common Corruptions}
\label{sec:prelim}

\begin{figure}[t]
    \centering
    \includegraphics[width=\textwidth]{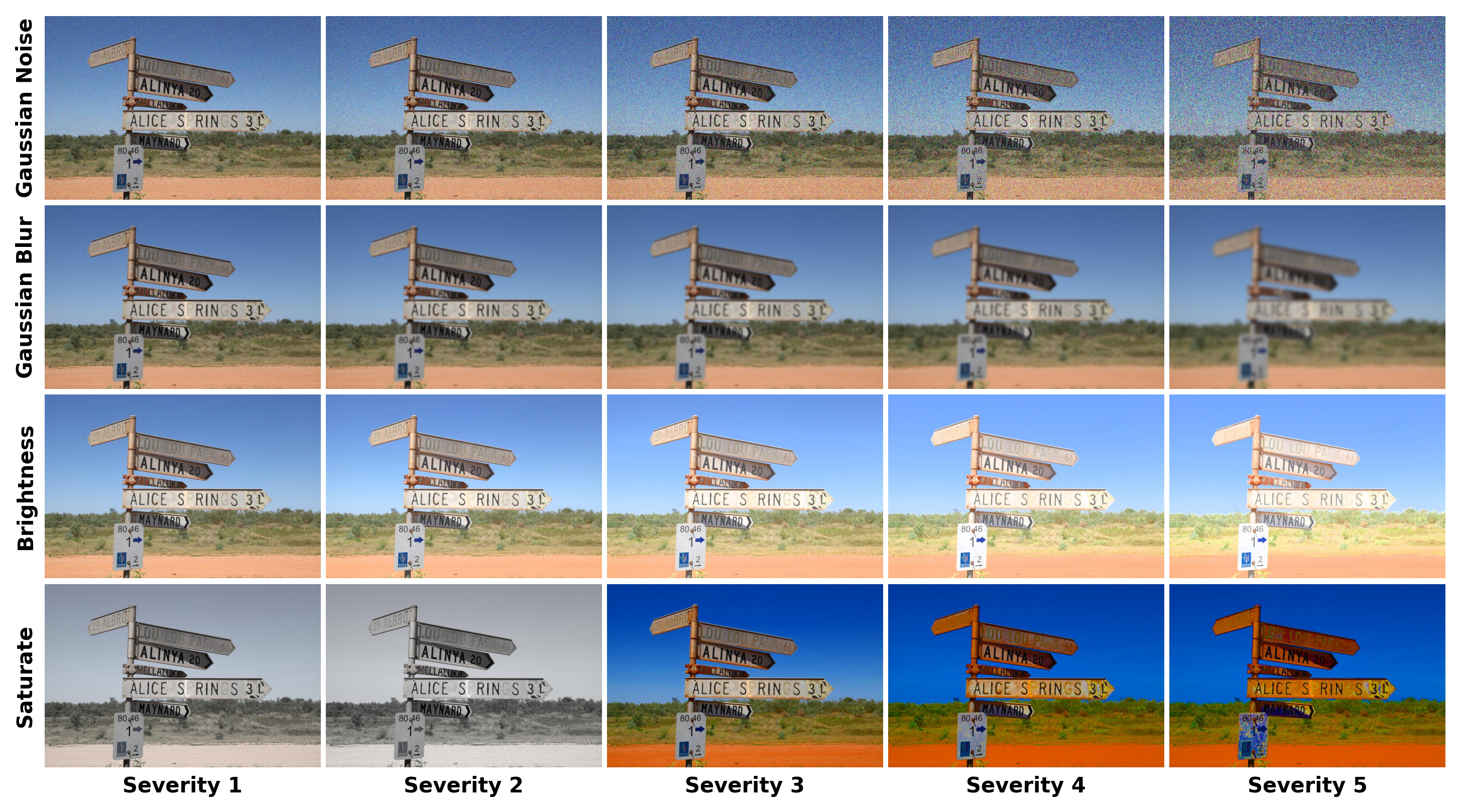}
    \caption{Different noise types across severity levels}
    \label{fig:noise-types}
\end{figure}

In this section, we provide background information on the 19 corruptions (in four categories) and implemented using the ImageNet-C framework~\cite{hendrycks2019benchmarking}, as well as their five severity levels. We show sample images across all severity levels for one corruption in each category in \autoref{fig:noise-types}.

\subsection{Noise Corruptions}

\textbf{Gaussian Noise} is an additive corruption that follows a normal (Gaussian) distribution, with noise values drawn from a Gaussian curve and applied uniformly across all pixels regardless of intensity. This makes it particularly relevant for modeling sensor-based electronic interference or ambient thermal fluctuations. It appears as small fluctuations in brightness or color and commonly occurs during image acquisition and transmission. Because it affects the entire image uniformly, Gaussian noise reduces overall image sharpness and degrades fine textures. It is extensively studied in image denoising tasks and used in the evaluation of restoration algorithms, particularly in photography, remote sensing, and video transmission systems.

\textbf{Shot noise} is an additive noise following a Poisson distribution, arising from the quantized nature of light; each pixel's noise is proportional to the square root of its intensity. This makes shot noise especially prominent in dark areas where fewer photons are captured, creating a characteristic grainy appearance. It is prevalent in low-light photography, astronomical imaging, medical X-ray systems, and night vision applications. Shot noise is inherent in photon-limited environments and poses significant challenges for precise pixel-based analysis.

\textbf{Impulse noise} is a sparse and discrete corruption following a binary or discrete distribution, typically manifesting as random black or white pixels scattered across the image. It represents abrupt disruptions, based on the impulse function, often caused by faulty sensors, memory cell errors, or communication dropouts. Because the noise is high in intensity but localized, it severely degrades image quality, particularly in fine textures and edges. This noise type is unpredictable and difficult to remove without degrading surrounding details. It is frequently used to evaluate the robustness of denoising and inpainting algorithms and is common in fault-tolerant systems and error-resilient transmission applications.

\textbf{Speckle noise} is a multiplicative corruption that typically follows a gamma or uniform distribution, where noise intensity scales with pixel intensity. Speckle noise is commonly found in coherent imaging systems like medical ultrasound, radar, and laser imaging. Its multiplicative nature causes it to obscure fine details more significantly in high-intensity areas, reducing contrast and image fidelity. Speckle noise plays a key role in the evaluation of medical imaging algorithms and remote sensing techniques, especially in applications like synthetic aperture radar (SAR) and optical coherence tomography (OCT).

\subsection{Blur Corruptions}

\textbf{Defocus Blur} is a convolutional corruption resulting from a deterministic circular point spread function (PSF), simulating a lens being out of focus. This blur appears uniform within a circular region and causes edges and fine structures to appear soft and smeared. Common causes include improper lens focus, shallow depth of field, and subject movement. Defocus blur significantly reduces spatial resolution and hampers feature-based vision systems. It is relevant in optical quality assessment and autofocusing algorithm research, as well as in applications such as mobile photography and biometric recognition.

\textbf{Glass Blur} simulates local pixel displacement using a random uniform distribution, imitating the scattering of light through a translucent surface. Pixels are locally shuffled within a neighborhood, distorting structural features. This corruption mimics the effects of privacy glass, foggy or dirty lenses, and is especially challenging because it disrupts both texture and shape information simultaneously. It is used in privacy-preserving vision systems and adversarial robustness evaluations, such as assessing model resilience to spatial perturbations.

\textbf{Motion Blur} is a convolutional corruption driven by a deterministic linear PSF. It simulates the smearing that occurs when the camera or subject moves during image capture. The linear blur direction and magnitude are often fixed or known, making it a structured corruption. It is commonly studied in deblurring and motion estimation research and is a frequent artifact in surveillance systems and sports video analytics.

\textbf{Zoom Blur} arises from a blend of frames or a convolutional radial PSF, simulating zoom-in or zoom-out motion during image capture. The radial distortion introduces a motion-like blur centered on the image's focal point. It models cinematic effects or accidental zooming, causing distortions that radiate outward and reduce detail, particularly in the center. This type of blur is relevant in video stabilization, cinematography simulation, and visual quality assessment in dynamic environments such as drone or bodycam footage.

\subsection{Weather Corruptions}

\textbf{Snow} is modeled as an additive overlay with procedural patterns, often generated using a uniform distribution to simulate snowflakes of varying sizes and transparency. These white specks obscure underlying content and reduce contrast, mimicking real-world winter conditions. Snow corruption is critical for evaluating autonomous navigation and surveillance systems in adverse weather. It is used to train and evaluate vision models for robustness in harsh weather conditions and to develop weather-aware understanding algorithms.

\textbf{Frost} is simulated as an additive texture with semi-opaque fractal or patterned overlays, replicating crystallization on lenses. The distribution of the pattern is often procedural and localized, simulating real frost accumulation. Frost introduces both blur and occlusion effects, significantly distorting the image's structure and reducing visibility. It is especially relevant in vehicle-mounted vision systems and outdoor security cameras operating in winter or high-humidity environments.

\textbf{Brightness} introduces a constant additive shift across all pixels, uniformly increasing or decreasing intensity values. This deterministic operation simulates overexposure or underexposure due to lighting changes or sensor miscalibration. Extreme brightness variations can mask content and impact visibility. Brightness adjustment simulations are crucial in HDR imaging and illumination normalization techniques, often included in pre-processing pipelines for robust model performance.

\textbf{Fog} is modeled as an additive haze based on depth-dependent exponential or linear scattering. Light diffusion reduces contrast and visibility, especially in background elements. It mimics atmospheric conditions and is vital for evaluating the robustness of vision systems in autonomous driving, drone imaging, and surveillance. Fog simulation is used in datasets and algorithms for weather-invariant perception and is crucial for depth estimation tasks under limited visibility.

\subsection{Digital Corruptions}

\textbf{Saturate} is a deterministic transformation that alters the chromatic intensity of an image by scaling the saturation component within color space, typically using a uniform scaling factor. This adjustment does not follow a probability distribution but applies a fixed or parameterized change, making it a non-random digital corruption. The transformation exaggerates or suppresses color vividness while preserving hue and luminance, resulting in images that range from highly vibrant to nearly grayscale. Saturate corruption is common in post-processing pipelines, artistic filters, or camera sensor miscalibrations.

\textbf{Contrast} applies a multiplicative linear scaling to pixel values, causing an increase or decrease in the intensity difference between lighter and darker regions. This can result in exaggerated or flattened features, depending on the scaling factor. Such alterations reflect real-world issues such as glare, shadows, or inconsistent lighting and affect feature extraction performance. Contrast adjustments are used in training models for contrast-invariant recognition tasks and are crucial in medical imaging, remote sensing, and document analysis.

\textbf{Elastic Transform} introduces spatial warping through a displacement field drawn from a Gaussian distribution. This creates smooth, localized deformations that alter the shape and structure of objects. Originally developed for data augmentation, such distortions also mimic imaging artifacts from flexible surfaces or low-quality capture conditions. Elastic transformations are commonly used in handwritten digit recognition, face morphing tasks, and training data augmentation in deep learning.

\textbf{Pixelate} results from deterministic downsampling followed by upsampling, creating block-like artifacts. The image is effectively averaged over square regions, simulating low-resolution or bandwidth constraints. This reduces fine detail and disrupts pattern recognition models that rely on texture information. It is widely used to evaluate robustness in compressed video and low-quality surveillance footage and presents a common challenge in facial recognition from anonymized or censored data.

\textbf{JPEG Compression} introduces compression artifacts caused by the quantization of noise in the frequency domain. As JPEG compresses an image by discarding high-frequency components, visible block boundaries and ringing artifacts appear, especially at high compression levels. It mimics aggressive storage or transmission compression, distorting texture and edge integrity. JPEG corruptions are relevant in multimedia quality assessment, forensic analysis, and deep learning-based compression artifact reduction systems.

\subsection{Impact of Severity Levels}

Each corruption mentioned above has five severity levels. The five severity levels represent increasing intensities of corruption applied to visual inputs, simulating progressive real-world degradation scenarios. Severity level 1 corresponds to mild distortions with subtle visual impact, while level 5 represents extreme degradation causing substantial loss in visual quality. These carefully calibrated progressions enable comprehensive assessment of model robustness across a spectrum of corruptions, revealing how performance deteriorates as input quality worsens. By evaluating models at multiple severity levels, we gain deeper insights into their tolerance thresholds and adaptability to increasingly noisy or distorted data, which is crucial for real-world deployment in unpredictable environments.

Each corruption type follows specific severity scaling patterns: in the Noise category, higher severity levels increase the standard deviation of distributions (as in Gaussian noise where larger variations alter pixel colors more dramatically) or probability of affected pixels, resulting in progressively grainier images; Blur corruptions intensify through larger kernel sizes or displacement factors, leading to stronger image smearing and loss of edge details; Weather corruptions modulate opacity, density, and contrast (such as in Brightness, where increasing values in the HSV color space create progressively overexposed regions); while Digital corruptions adjust parameters like compression rates, pixel block sizes, or color transformations (as demonstrated in Saturate, where severity levels 1-2 decrease color vividness while levels 3-5 dramatically increase it, creating unnaturally vivid and distorted colors). Each severity function is meticulously tuned to ensure level 1 represents a mild disturbance that minimally impacts model performance, while level 5 poses a significant challenge even to human visual perception, establishing a comprehensive benchmark for evaluating vision-language model resilience.
\section{Evaluation Methodology}
\label{sec:method}

This section outlines our comprehensive approach to assessing Vision-Language Model robustness under varied corruption conditions. We first describe the LLaVA 1.5 architecture used in our experiments, then introduce our corruption-augmented datasets designed specifically for this evaluation, and finally detail the metrics employed to quantify performance degradation.

\subsection{LLaVA 1.5 Vision Language Model}

\begin{figure}[t]
    \centering
    \includegraphics[width=0.5\textwidth]{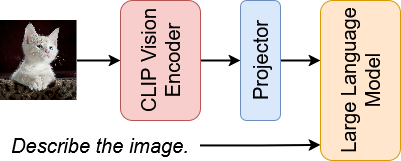}
    \caption{Overview of the LLaVA architecture}
    \label{fig:llava-architecture}
\end{figure}

LLaVA (Large Language and Vision Assistant) integrates a powerful vision encoder with a large language model (LLM) through a lightweight projection layer (as shown in \autoref{fig:llava-architecture}). This architecture enables effective multimodal instruction-following without requiring complex fusion mechanisms.

The visual input $X_v$ is first processed by the CLIP ViT-L/14~\cite{radford2021learning} encoder, denoted as $g(\cdot)$, to extract high-level image features $Z_v$:

\begin{equation}
Z_v = g(X_v)
\label{eq:vision-encoder}
\end{equation}

To bridge the gap between visual features and the language model's token embedding space, a trainable linear projection layer $W$ transforms $Z_v$ into token-like embeddings $H_v$:

\begin{equation}
H_v = W \cdot Z_v
\label{eq:projection}
\end{equation}

These projected visual tokens $H_v$ are treated equivalently to text tokens and are concatenated with the natural language input tokens before being passed to the LLM. LLaVA~\cite{liu2023llava} employs Vicuna~\cite{vicuna2023}, a fine-tuned version of LLaMA 2.0 7B~\cite{touvron2023llama}, as its language model backbone due to its strong instruction-following capabilities.

This design ensures that visual and textual modalities are jointly processed by the autoregressive decoder. Unlike more intricate fusion strategies such as the Perceiver Resampler in Flamingo~\cite{alayrac2022flamingo} or the Querying Transformer (Q-Former) in BLIP-2~\cite{li2023blip}, LLaVA~\cite{liu2023llava} prioritizes simplicity and efficiency, reserving more sophisticated integration techniques for future exploration.

Training LLaVA~\cite{liu2023llava} involves two sequential stages: an initial pretraining stage for multimodal alignment, followed by supervised fine-tuning for vision-language instruction tasks. The alignment phase focuses on training the projection layer (\autoref{eq:projection}) to enable the LLM to effectively interpret the visual embeddings produced in \autoref{eq:vision-encoder}. This process utilizes a curated instruction-following dataset, where each sample consists of an image, a prompt, and a language response.

This training objective guides the model to generate coherent, contextually relevant responses grounded in the image content. The use of GPT-4-generated~\cite{ai2023gpt} instruction data further enhances the model's instruction-following behavior and its ability to reason over visual content. For each image $\mathbf{x}_v$, the authors generate multi-turn conversation data 
$(\mathbf{X}^1_q, \mathbf{X}^1_a, \ldots, \mathbf{X}^T_q, \mathbf{X}^T_a)$, 
where $T$ denotes the total number of turns. These are organized sequentially by treating all answers as the assistant's responses, and defining the instruction $\mathbf{X}^t_{\text{instr}}$ at the $t$-th turn as:

\begin{equation}
\mathbf{X}^{t}_{\text{instr}} = 
\begin{cases} 
\text{Randomly choose } [\mathbf{X}^{1}_{q}, \mathbf{X}_{v}] \text{ or } [\mathbf{X}_{v}, \mathbf{X}^{1}_{q}], & \text{if } t = 1 \\
\mathbf{X}^{t}_{q}, & \text{if } t > 1
\end{cases}
\end{equation}

This formulation establishes a unified format for the multimodal instruction-following sequence. The model is instruction-tuned on the prediction tokens using the original auto-regressive training objective of the language model.

Specifically, for a sequence of length $L$, the model computes the probability of the target answers $\mathbf{X}_a$ as:

\begin{equation}
p(\mathbf{X}_a \mid \mathbf{X}_v, \mathbf{X}_{\text{instr}}) = 
\prod_{t=1}^{L} p(X_{a}^{t} \mid \mathbf{X}_v, \mathbf{X}_{\text{instr}}, <t, \mathbf{X}_a)
\end{equation}

At inference time, given a user prompt and an image, the model executes the following steps: first, the image $X_v$ is encoded into visual features $Z_v$ using \autoref{eq:vision-encoder}. Then, these features are linearly projected into token embeddings $H_v$ via \autoref{eq:projection}. The projected embeddings $H_v$ are prepended to the tokenized user prompt. Finally, this sequence is passed to the LLM, which generates output tokens in an autoregressive fashion.

This process enables LLaVA~\cite{liu2023llava} to respond to image-grounded natural language instructions in a fluent and informative manner, demonstrating capabilities across a range of vision-language tasks including captioning, visual question answering, and visual dialogue.

\subsection{Corruption-Augmented Benchmark Datasets}

Despite the growing importance of Vision-Language Models in real-world applications, there exists a notable gap in standardized benchmarks for evaluating their robustness under various corruption conditions. Drawing inspiration from the ImageNet-C framework~\cite{hendrycks2019benchmarking}, we introduce TextVQA-C and GQA-C, two novel corruption-augmented datasets specifically designed to assess VLM resilience across diverse visual degradation scenarios.

We selected TextVQA and GQA as our foundation datasets due to their complementary focus areas, providing a comprehensive evaluation across different visual reasoning capabilities. TextVQA specifically targets the model's ability to read and reason about textual information within images, making it particularly valuable for assessing how corruptions affect text recognition capabilities. GQA, conversely, emphasizes spatial reasoning and understanding of object relationships within images, offering insights into how corruptions impact more general visual comprehension tasks (refer to samples of both datasets in \autoref{fig:vqa_examples}).

\begin{figure}[t]
    \begin{subfigure}{0.45\textwidth}
        \centering
        \includegraphics[width=0.6\linewidth]{Images/TextVQA_example_Image.jpg}
        \caption{TextVQA Question: \textit{What is the year on the calender?} Answer: \textit{2013}.}
        \label{fig:TextVQA Example}
    \end{subfigure}
    \hfill
    \begin{subfigure}{0.45\textwidth}
        \centering
        \includegraphics[width=0.6\linewidth]{Images/GQA_example_image.jpg}
        \caption{GQA Question: \textit{Is it overcast?} Answer: \textit{No}.}
        \label{fig:GQA Example}
    \end{subfigure}
    \caption{Sample image-text pairs from TextVQA and GQA.}
    \label{fig:vqa_examples}
\end{figure}

For our evaluation, we utilized 5,000 samples from TextVQA and 12,578 samples from GQA, applying the full suite of 19 corruptions at 5 severity levels to each image. Both datasets employ concise, typically single-word answers, which facilitates precise evaluation and reduces ambiguity in model performance assessment. This design choice is particularly important when evaluating performance degradation under corruption, as it eliminates variability in answer phrasing that might confound the analysis of corruption effects.

\subsection{Performance Metrics}

We evaluate LLaVA 1.5's performance on both datasets using accuracy metrics tailored to their specific characteristics. For TextVQA, which focuses on reading and understanding text in images, the model processes the image and question, then generates a natural language answer. This answer is evaluated against a set of human-annotated ground truths, with accuracy determined by whether the prediction matches at least one of the acceptable answers, accommodating natural variations in human responses.

For GQA, which tests spatial reasoning and visual understanding, LLaVA 1.5 generates concise answers that are compared directly to single ground truth responses, employing a more stringent evaluation criterion. The differential impact of corruptions on these datasets provides valuable insights: TextVQA performance is particularly vulnerable to corruptions that affect text visibility and clarity, while GQA performance is more sensitive to corruptions that degrade overall image quality and object recognition. In both cases, the model's ability to integrate corrupted visual features with textual context determines its robustness under challenging visual conditions.
\section{Results and Analysis}
\label{sec:analysis}

Across the four categories of corruption, the baseline clean accuracy for GQA proves to be higher than that of TextVQA, at 62.5\% and ~57.5\% respectively. The trends across the four categories reveal key differences in how transformers handle different corruptions at the five severity levels, as well as the rate of decline in accuracies across the four corruption categories. TextVQA focuses on images containing textual elements, requiring models to read and understand text within the image to answer questions. GQA centers on object-based images, testing reasoning about objects, their attributes, spatial relations, and colors. TextVQA challenges visual-textual understanding, combining OCR with visual context, while GQA emphasizes compositional reasoning and scene understanding beyond surface-level object detection. We further discuss trends within individual categories in the following subsections.

\begin{figure}[t]
    \centering
    \includegraphics[trim=0 0 0 32 , clip, width=\textwidth]{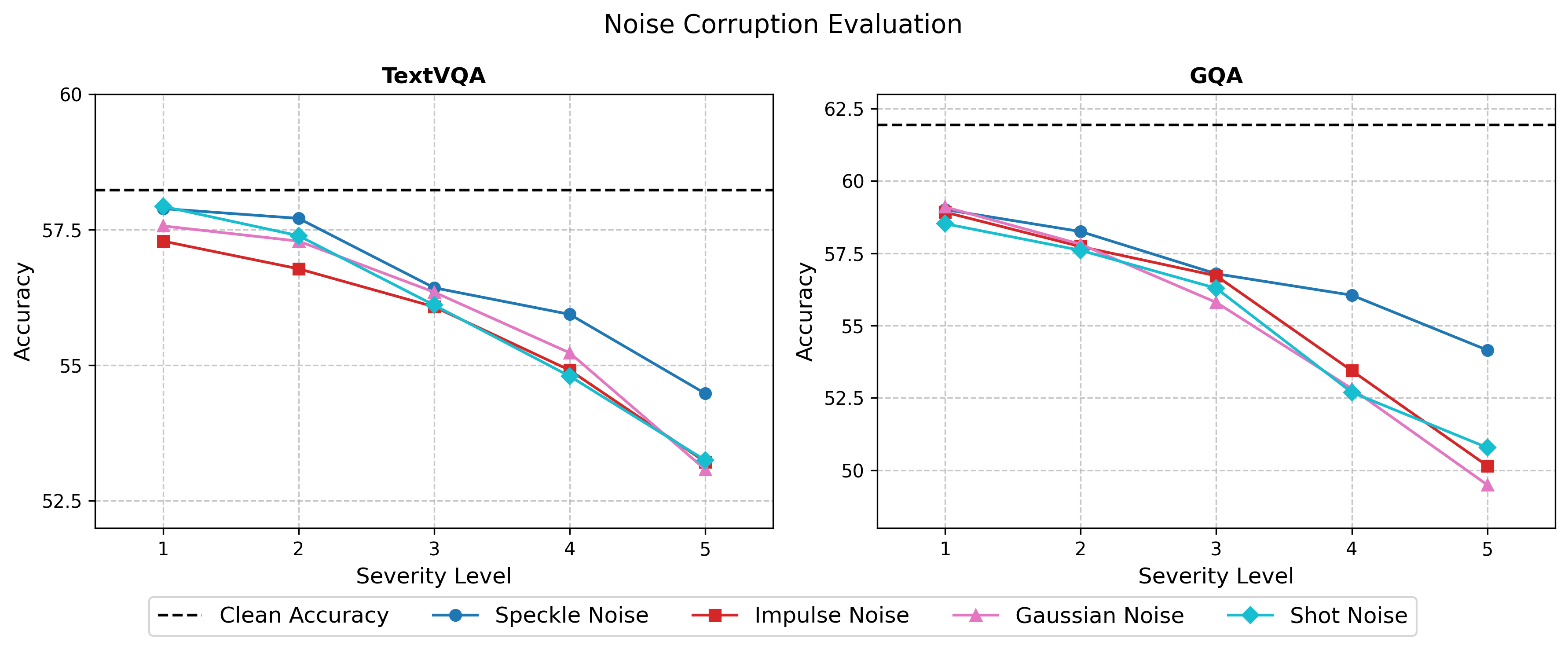}
    \caption{Accuracy Results for the Noise Corruption Category}
    \label{fig:noise-corruption}
\end{figure}

\subsection{Noise Corruptions}
The noise corruption trends (shown in \autoref{fig:noise-corruption}) in the graph reveal key differences in how transformers handle visual noise across tasks. In both TextVQA and GQA, accuracy steadily declines with increased noise severity, but the rate and pattern of this decline vary across the four noise types.

Speckle noise shows the slowest degradation across both datasets, maintaining higher accuracy at each severity level—especially at severity 5, where TextVQA retains 54\% and GQA stays above 54\%. This indicates that the LLaVA 1.5 model is most robust to speckle noise, compared to the other three noises in the category.

In contrast, impulse noise causes the steepest collapse in both TextVQA and GQA, particularly visible from severity 3 onward, where accuracy sharply drops to around 53\% for TextVQA and dips to 50\% for GQA by severity 5. Initially, impulse noise has the lowest accuracy in TextVQA but starts at approximately the same accuracy as the other noises in GQA.

Gaussian noise demonstrates a smoother decline, moderately affecting performance, although having the lowest accuracy below 50\% in GQA (from 57\% to 53\% in TextVQA and a similar steeper curve in GQA). Overall, Gaussian noise follows roughly the same trend in both TextVQA and GQA, achieving a lower final accuracy in GQA compared to TextVQA.

Shot noise presents an initially strong accuracy (57\% in TextVQA at severity 1) followed by a sharp fall—highlighting intensity-dependent disruption patterns. In GQA, shot noise demonstrates a more shallow decline after severity 4, unlike impulse and gaussian noise that show steeper declines.

\textbf{Frequency-Domain Perspective:} The observed robustness patterns align with transformers' inherent frequency processing characteristics. Transformer-based vision encoders like CLIP's ViT employ self-attention mechanisms that act as weighted averages, naturally emphasizing low-frequency components while attenuating high-frequency details. This explains why speckle noise, which primarily preserves low-frequency structures, causes the least degradation across both datasets. Conversely, impulse noise introduces sharp high-frequency disruptions that directly conflict with the transformer's processing bias, resulting in the steepest performance drops. The relative resilience of TextVQA to certain noise types, despite its OCR-dependency, suggests that transformer models can leverage broader contextual understanding to compensate for local text degradation. This frequency-selective behavior differentiates transformers from CNNs, which typically exhibit stronger sensitivity to low-frequency corruptions while better preserving high-frequency details.

\subsection{Blur Corruptions}

\begin{figure}[t]
    \centering
    \includegraphics[trim=0 0 0 32, clip, width=\textwidth]{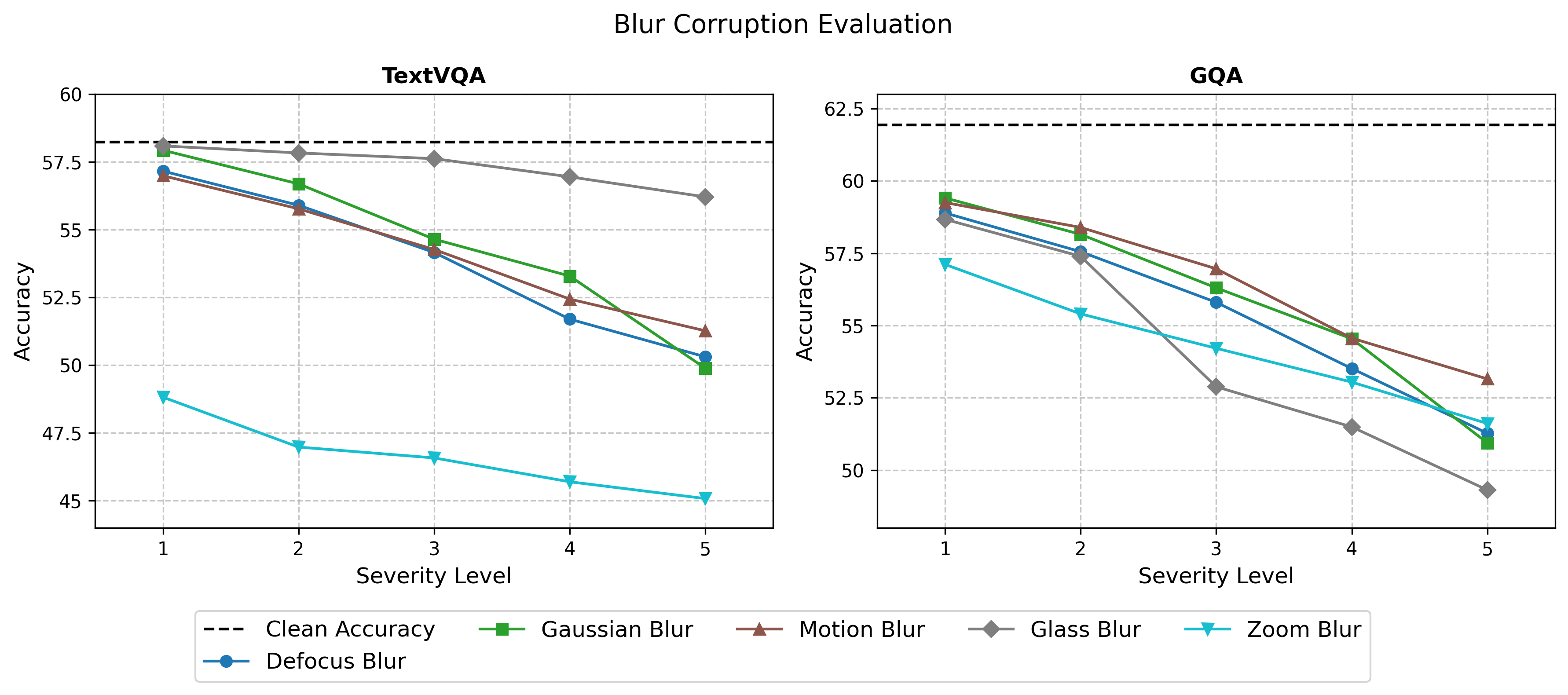}
    \caption{Accuracy Results for the Blur Corruption Category}
    \label{fig:blur-corruption}
\end{figure}

In the Blur Corruption Evaluation (presented in \autoref{fig:blur-corruption}), the performance trends of the LLaVA 1.5 model across different blur types show clear distinctions in robustness, particularly when comparing the TextVQA and GQA tasks. Across both tasks, the model's accuracy declines steadily with increased blur severity, but the impact varies: TextVQA generally retains performance longer, while GQA is more sensitive to precise spatial degradation. 

In the TextVQA plot, LLaVA 1.5 emerges as the most robust to Glass Blur, maintaining the highest accuracy across all severity levels and ending with 56\% accuracy at severity 5. This suggests that despite its distortion, Glass Blur preserves enough structural information for the model to extract useful context. Glass Blur, the strongest performer in TextVQA, shows the steepest decline in GQA, decreasing from 58\% at severity 1 to just below 50\% at severity 5. This difference reflects the nature of each dataset.

Motion Blur shows moderate resilience in both datasets, declining gradually and converging around 51–52\% at severity 5. In GQA, which is more reliant on precise visual detail for object recognition and spatial relationships, the robustness order shifts slightly. Motion Blur is the most resistant to corruption here, ending near 53\% at severity 5, likely due to its relatively predictable directional degradation. 

Defocus blur is observed to have approximately a 7-point drop in both TextVQA and GQA. In TextVQA, the accuracy follows a trajectory similar to that of Motion Blur initially, then dips more steeply after severity level 3, before finishing at its final accuracy of around 50\%. In GQA, the decline remains steep after severity 3, compared to a less steep decline after severity level 4 in TextVQA.

Zoom Blur stands out as the least robust in TextVQA, with a steep and consistent decline from nearly 49\% at severity 1 to just 45\% at severity 5—highlighting the model's sensitivity to global distortion. Zoom Blur shows better robustness in GQA than in TextVQA, maintaining above 51\% accuracy at severity 5, while in TextVQA, the baseline of zoom blur began lower than 50\%.

Gaussian Blur, initially having the highest accuracy in both TextVQA and GQA, finished at an accuracy lower than the other three, excluding zoom blur. The initial highest accuracy is observed in GQA rather than TextVQA, despite having the higher initial accuracy in both datasets. The overall trajectory for both TextVQA and GQA seems to have a steeper dip after severity level 4.

\textbf{Frequency-Domain Perspective:} The varying effects of blur corruptions can be understood through the lens of frequency domain processing in transformer architectures. Blur operations fundamentally act as low-pass filters, attenuating high-frequency information while preserving low-frequency content to different degrees. Transformer-based models like LLaVA 1.5 naturally emphasize low-frequency components through their self-attention mechanisms, making them relatively resilient to certain blur types. Glass blur, despite introducing local distortions, preserves global low-frequency structures that transformers can leverage, explaining its minimal impact on TextVQA. Conversely, Gaussian blur's uniform suppression of high frequencies significantly impairs fine-grained object recognition in GQA tasks, where spatial relationships are crucial. The relatively better performance of Motion blur in GQA suggests that directional blurs preserve more discriminative information along certain axes, allowing the transformer to still capture relevant object relationships despite the degradation. This frequency-selective resilience underscores how the inherent inductive biases of transformer architectures interact with different types of signal degradation.

\begin{figure}[t]
    \centering
    \includegraphics[trim=0 0 0 32, clip, width=\textwidth]{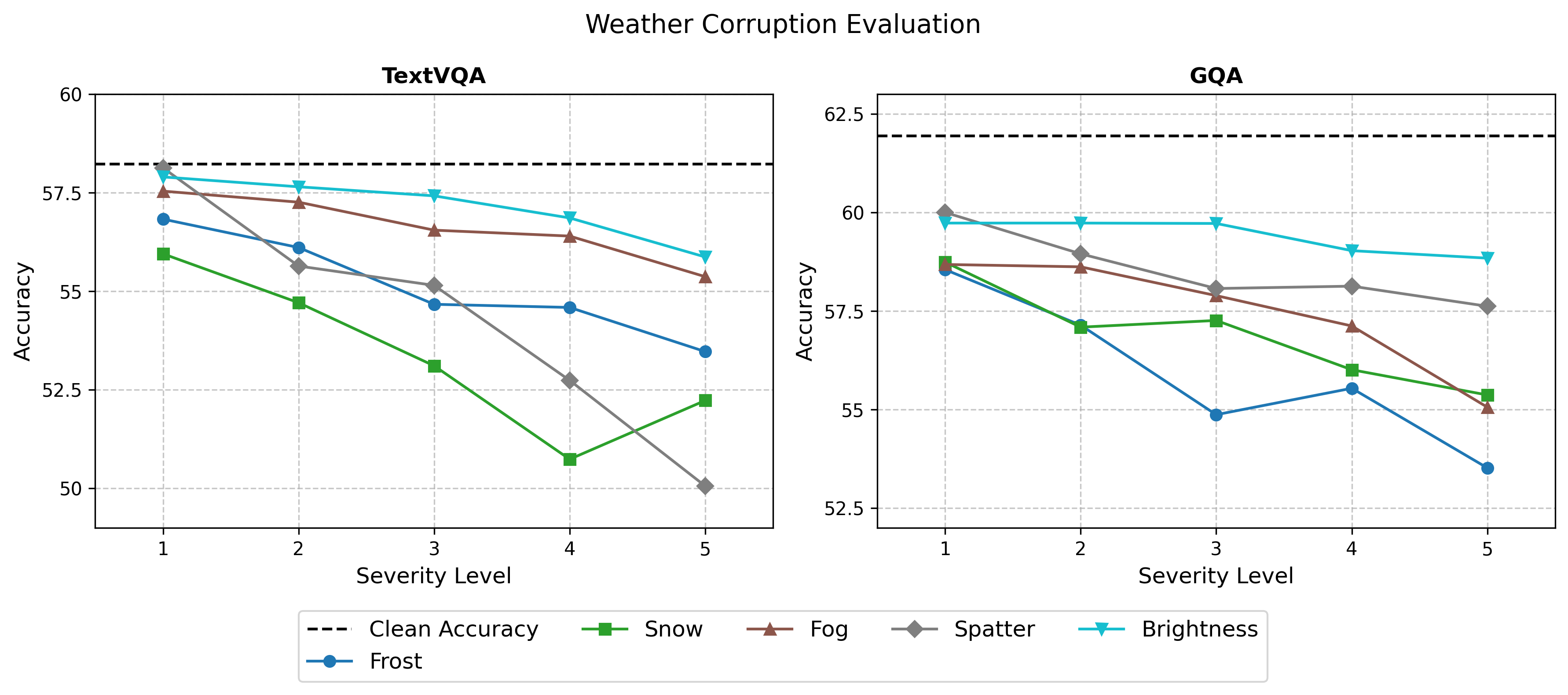}
    \caption{Accuracy Results for the Weather Corruption Category}
    \label{fig:weather-corruption}
\end{figure}

\subsection{Weather Corruptions}

In the Weather Corruption Evaluation (shown in \autoref{fig:weather-corruption}), the performance trends of the LLaVA 1.5 model across different weather types show clear distinctions in robustness. Across both tasks, the model's accuracy declines steadily with increased weather severity, but the impact varies. Overall, it is observed that TextVQA generally performs worse compared to GQA as textual information is more likely to lose information when images are altered either from an increase in brightness or artifacts being introduced from weather such as fog, snow, and frost. For GQA, the overall accuracy drop is less steep, suggesting better robustness.

Frost shows a less steep but still notable drop, reaching an accuracy of 53\% at severity 5. Meanwhile, fog demonstrates the highest resilience among natural weather types. In GQA, Frost leads to the sharpest decline, dropping by almost 4 percentage points. Frost performed worse for the GQA dataset as it affects the object's details, causing the model to lose visual information from the image, making it difficult to identify the key characteristics of an image. Frost shows an anomalous value at severity 4 in GQA evaluation where the accuracy rises from severity 3 to 4 and falls back down to about 53\% in severity 5.

Brightness causes the smallest degradation, with accuracy decreasing marginally, indicating that brightness changes have minimal impact on TextVQA performance. In GQA, brightness remains the most stable, with accuracy only slightly decreasing. We also see that brightness has a lesser impact on the model's accuracy across all datasets. This indicates that VLM models are better at extracting information in light environments compared to contrast where the model performs worse.

Snow is observed to have an initially higher starting accuracy in GQA than TextVQA, with the decline being the steepest after severity 4 in TextVQA, before the only rise observed from severity 4 to 5. In TextVQA, the most severe drop is observed with snow, the steepest among all weather types. Snow also shows an anomalous value at severity 5, with the accuracy decreasing before rising again from severity 4 to 5.

Spatter observes a different trajectory for both datasets. In TextVQA, although initially having the highest accuracy, it steeply declines to severity level 2, and then declines steeply once again after severity level 3. In contrast, the accuracy is once again the highest in GQA initially, but the trajectory is not as steep as that in TextVQA. Only an approximately 3-point difference is shown between the initial and final accuracy for spatter.

The model is shown to be least robust to snow and spatter corruptions. These types are similar as they are grainy corruptions. Introducing grain particularly affects textual information the most, as this can be seen when we introduced impulse and gaussian noise which introduced grain-like artifacts and had affected the TextVQA accuracy percentage considerably. Based on a comparison between the two plots for the datasets, the model is seen to be most robust to brightness.

\textbf{Frequency-Domain Perspective:} Weather corruptions affect image frequency content in distinct ways that interact meaningfully with transformer processing characteristics. Brightness changes primarily modify image intensity while largely preserving frequency structures, explaining the minimal impact on both datasets—transformers can still effectively extract the same spatial relationships and textual cues from the preserved frequency patterns. In contrast, snow and spatter introduce high-frequency noise components that disrupt the transformer's preferred low-frequency processing, particularly affecting TextVQA where fine-grained text features are critical. Fog and frost operate differently; fog applies a low-pass filtering effect by reducing contrast and adding uniform intensity, while frost introduces structured high-frequency patterns. The transformer's self-attention mechanism, which emphasizes global context through weighted averaging, can partly compensate for fog's uniform degradation but struggles with frost's structured disruptions that interfere with patch-level feature extraction. This frequency-domain perspective explains why weather corruptions that preserve low-frequency structure (brightness, fog) affect performance less than those introducing competing high-frequency patterns (snow, spatter, frost).

\begin{figure}[t]
    \centering
    \includegraphics[trim=0 0 0 32, clip, width=\textwidth]{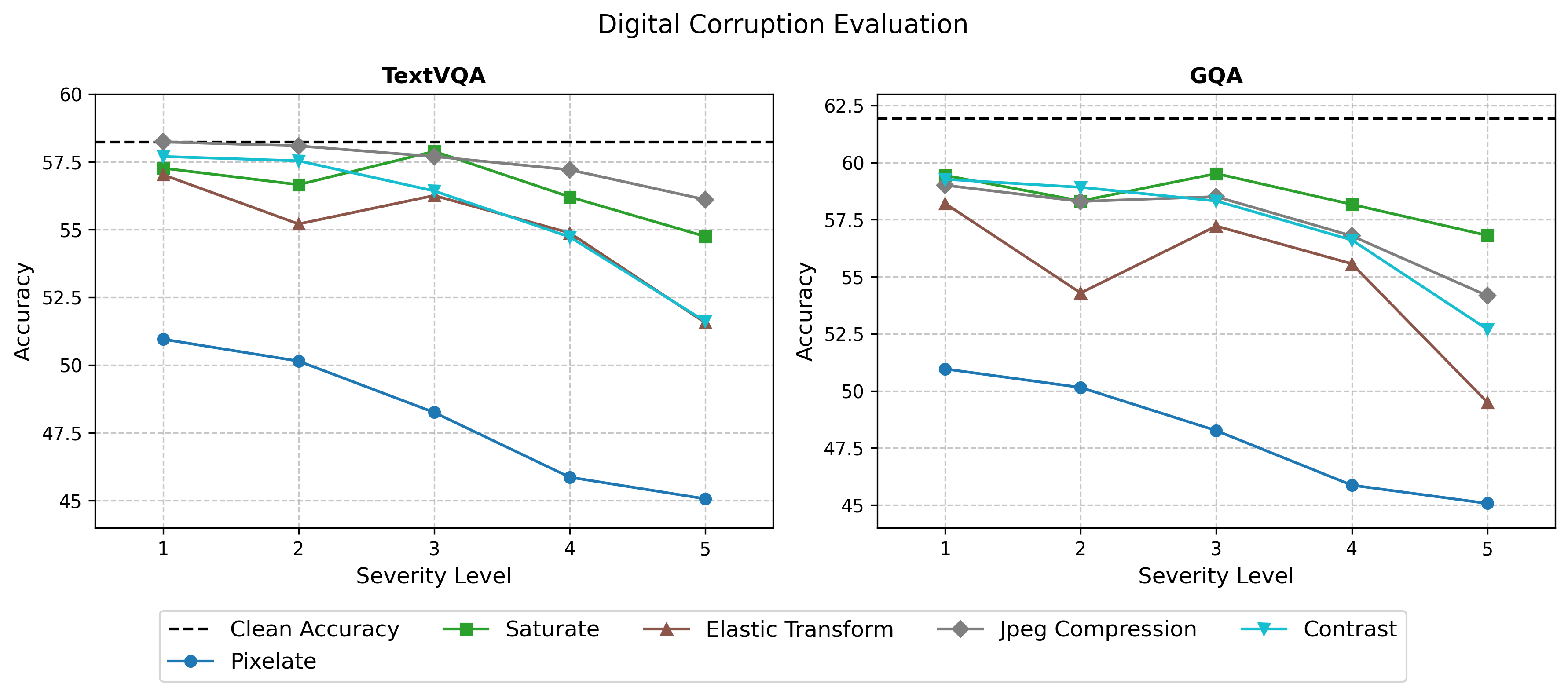}
    \caption{Accuracy Results for the Digital Corruption Category}
    \label{fig:digital-corruption}
\end{figure}

\subsection{Digital Corruptions}

Across the digital category (results presented in \autoref{fig:digital-corruption}), we observe that Pixelate introduces a similar accuracy degradation in both GQA and TextVQA datasets, with accuracy consistently dropping from 51\% to 45\% as severity increases. This suggests that pixelation equally degrades the visual quality of object features (like shape, color, and spatial relationships in GQA) and textual elements (as required in TextVQA). It indicates that Pixelate is a broadly effective corruption, impairing both object recognition and text recognition capabilities in visual question answering tasks.

Contrast adjustments, on the other hand, do not significantly impact GQA accuracy (ranging from 59\% to 52.7\%), indicating that object features like shapes and spatial relationships remain largely unaffected. However, in TextVQA, contrast changes cause a more noticeable drop in accuracy (from 57.4\% to 51.6\%), showing that text recognition is more sensitive to contrast variations and reliant on clear edge and font visibility.

Elastic Transform leads to a greater accuracy drop in GQA, suggesting it effectively distorts object shapes and layouts that are essential for scene understanding. The accuracy drops from ~58.4\% to ~49.3\% across severity levels. In TextVQA, Elastic Transform and Contrast cause similar levels of degradation (falling to ~51.6\%), implying that both are equally effective at disrupting text-based visual information, though in different ways—contrast affects visibility, while elastic transform affects spatial coherence.

Additionally, JPEG Compression and Saturation result in relatively smaller accuracy drops, indicating that models are somewhat robust to mild digital distortions. GQA and TextVQA show only a minor decline in performance under these corruptions, especially at lower severity levels.

\textbf{Frequency-Domain Perspective:} Digital corruptions provide further insight into transformer models' frequency processing characteristics. Pixelation fundamentally decimates high-frequency information while introducing artificial high-frequency edges at pixel boundaries—a dual disruption that severely impacts transformer processing. This explains the consistent degradation across both datasets, as pixelation disrupts both the global scene structure and local details needed for text recognition. JPEG compression, which primarily removes high-frequency components while preserving low-frequency structures, causes minimal degradation in line with transformers' low-frequency processing bias. Elastic transforms are particularly interesting as they warp the spatial relationships between image regions without significantly altering frequency content within those regions—directly challenging the spatial correspondence assumptions in transformer self-attention mechanisms. The differential impact of contrast adjustments between TextVQA and GQA reflects how contrast primarily affects edge definition (high-frequency components) critical for text recognition, while transformers can still extract object relationships from the preserved low-frequency structural information. These observations highlight how transformer models' patch-based processing and attention mechanisms interact with different frequency disruptions, with corruptions that preserve global structure and low-frequency information causing less performance degradation.

\subsection{Overall Analysis}
\textbf{TextVQA:} In the TextVQA dataset, the most prominent trend is the drastic performance degradation under \textbf{blur corruption}, with Defocus Blur and Glass Blur causing the fastest and most severe accuracy degradation among all types. These distortions significantly reduce the clarity of embedded text, making it nearly unreadable for the model's OCR and visual-textual reasoning components. The graph clearly shows a steep downward curve for these corruptions as severity increases. \textbf{Weather corruptions} follow a similar trend, with Snow in particular leading to the worst performance among all weather types, due to its strong occlusion of scene elements and text. In contrast, Fog shows a relatively smoother degradation, possibly because the model retains enough contextual information for partial reasoning. \textbf{Noise corruptions}, such as Gaussian and impulse noise, produce moderate degradation, mostly by interfering with OCR-like components. Impulse noise leads to a particularly sharp degradation. \textbf{Digital Corruptions} like JPEG compression and Contrast changes cause less accuracy drop even at higher severity levels. However, Pixelate and Elastic Transform still lead to a noticeable decline, likely by distorting text shapes. An interesting behavior is observed in how the model handles JPEG compression, where performance remains relatively stable even under severe distortion. Overall, TextVQA models are most vulnerable to blur and snow-based weather corruptions, and most robust under digital compression or low-level Gaussian noise, with degradation patterns often aligning with the degree to which scene text remains visually accessible.

\textbf{GQA:} In the GQA dataset, corruption impacts model performance differently than in TextVQA, primarily due to GQA's reliance on object recognition, spatial reasoning, and color features. \textbf{Noise Corruption}: Impulse noise triggers the sharpest performance degradation, with accuracy dropping to around 50\% at severity 5, while Speckle noise proves the most tolerable, maintaining stability at higher severities. Shot noise and Gaussian noise cause moderate declines, but shot noise's impact flattens after severity 4. \textbf{Blur Corruption}: Motion blur shows the best resilience, likely due to its directional consistency that preserves overall object shape, while Glass Blur causes the steepest drop—its fragmenting effect distorts essential object-level visual cues. Zoom Blur, although harmful in TextVQA, is better tolerated in GQA, maintaining performance above 51\% even at severity 5. \textbf{Weather Corruptions} also reveal interesting patterns. Frost degrades performance the most, likely due to its overlay pattern obscuring object details, while Brightness and Spatter have negligible effects, indicating the model's strong adaptability to lighting changes and minor occlusions. \textbf{Digital corruptions}: Elastic Transform and Pixelate corruptions are most harmful, distorting spatial coherence and object outlines, whereas JPEG compression, Saturation, and Contrast barely dent the accuracy, reflecting the model's robustness against digital post-processing artifacts. Overall, GQA shows the highest resilience to Motion blur, Brightness, and JPEG distortions, and is most vulnerable to Frost, Impulse Noise, and Glass Blur, with degradation tied closely to how corruptions affect the fine details critical for object-level reasoning.

\textbf{Frequency-Domain Perspective:} From a frequency-domain perspective, the differential impact of corruptions across TextVQA and GQA aligns with transformer models' inherent processing biases. Transformer-based vision encoders, through their self-attention mechanisms, naturally emphasize low-frequency, global patterns while being less sensitive to high-frequency details. This explains why corruptions that primarily disrupt high-frequency components (JPEG compression, brightness changes) generally cause minimal degradation across both datasets. Conversely, corruptions that introduce competing high-frequency patterns (impulse noise, frost) or severely attenuate mid-frequency structures critical for object recognition (glass blur, elastic transform) show the most pronounced negative effects. The markedly different impacts between TextVQA and GQA for certain corruptions (e.g., zoom blur, glass blur) highlight how text recognition requires preservation of specific mid-to-high frequency bands for character edges, while object recognition in GQA can leverage lower-frequency shape information. This frequency-selective vulnerability provides valuable insights for developing more robust vision-language models, suggesting that architectural innovations should address the transformer's inherent bias toward low-frequency processing when fine-grained visual details are critical for the task.

\section{Conclusion}
\label{sec:conc}

Our systematic analysis of VLM robustness across 19 corruption types reveals distinct vulnerability patterns in text and object understanding tasks. We find that transformer-based models like LLaVA 1.5 demonstrate an inherent bias toward low-frequency information processing, making them more resilient to corruptions that preserve global structure while being particularly vulnerable to high-frequency disruptions. Text recognition tasks prove especially sensitive to blur and snow corruptions, while object reasoning shows higher vulnerability to frost and impulse noise. These findings emphasize the importance of considering both task-specific requirements and architectural characteristics when designing robust vision-language models, suggesting that future architectures should address frequency-selective vulnerabilities, particularly for applications requiring fine-grained visual understanding.

\clearpage

\bibliographystyle{ieeetr}
\bibliography{bls_final}

\section*{Authors}

\noindent {\bf M. Usama} is pursuing a Bachelor's degree in Electrical Engineering from DHA Suffa University, Karachi, Pakistan. His research interests include Cybersecurity (OT security), Multimodal AI, Computer networks, and Computer Vision.\\

\noindent {\bf S. A. Asim} is pursuing a Bachelor's degree in Electrical Engineering from DHA Suffa University, Karachi, Pakistan. Her research interests include Computer Vision, Generative AI, Machine Learning, and Ethical AI.\\

\noindent {\bf S. B. Ali} is pursuing a Bachelor's degree in Electrical Engineering from DHA Suffa University, Karachi, Pakistan. His research interests include Embedded Systems Programming, Image and Signal Processing.\\

\noindent {\bf S. T. Wasim} is a Ph.D. student at the University of Bonn, Germany with a focus on Multimodal Vision Language Models.\\

\noindent {\bf U. B. Mansoor} works as Assistant Professor in Electrical Engineering at DHA Suffa University, Karachi, Pakistan. His research interests include Signal Processing, Machine Learning, and Large Language Models.\\

\end{document}